\documentclass{article}
\usepackage{spconf,amsmath,graphicx}
\usepackage{comment}

\usepackage[pagebackref,breaklinks,colorlinks]{hyperref}

\usepackage{soul}
\usepackage{graphicx}
\usepackage{amsmath}
\usepackage{amssymb}
\usepackage{booktabs}
\usepackage{comment}
\usepackage{multirow}
\usepackage{float}

\title{FootBots: A Transformer-based Architecture\\ for Motion Prediction in Soccer}
\name{Guillem Capellera$^{1, 2}$ \quad Luis Ferraz$^{1}$ \quad Antonio Rubio$^{1}$ \quad Antonio Agudo$^{2}$ \quad Francesc Moreno-Noguer$^{2}$\thanks{This work has been supported by the project MoHuCo PID2020-120049RB-I00 funded by MCIN/AEI/10.13039/501100011033 and by the Government of Catalonia under 2023 DI 00058.}}

\usepackage{fancyhdr}
\address{$^{1}$ Kognia Sports Intelligence \qquad $^{2}$ Institut de Robòtica i Informàtica Industrial, CSIC-UPC}

\begin{document}
%

\maketitle

\begin{abstract}
Motion prediction in soccer involves capturing complex dynamics from player and ball interactions. We present FootBots, an encoder-decoder transformer-based architecture addressing motion prediction and conditioned motion prediction through equivariance properties. FootBots captures temporal and social dynamics using set attention blocks and multi-attention block decoder. Our evaluation utilizes two datasets: a real soccer dataset and a tailored synthetic one. Insights from the synthetic dataset highlight the effectiveness of FootBots' social attention mechanism and the significance of conditioned motion prediction. Empirical results on real soccer data demonstrate that FootBots outperforms baselines in motion prediction and excels in conditioned tasks, such as predicting the players based on the ball position, predicting the offensive (defensive) team based on the ball and the defensive (offensive) team, and predicting the ball position based on all players. Our evaluation connects quantitative and qualitative findings. \href{https://youtu.be/9kaEkfzG3L8}{https://youtu.be/9kaEkfzG3L8}
\end{abstract}
\begin{keywords}
Motion prediction, Signal forecasting, Transformer, Trajectory understanding, Soccer.
\end{keywords}

\section{Introduction}
\label{sec:intro}

Multi-agent Motion Prediction (MP) holds critical importance across various domains, encompassing financial economics~\cite{sezer2020financial}, human pose estimation~\cite{fragkiadaki2015recurrent,martinez2017human,aksan2021spatio,guo2023back}, pedestrian behavior analysis~\cite{alahi2016social,gupta2018social,kosaraju2019social,salzmann2020trajectron++,ngiam2021scene,girgis2021latent}, and sports analytics~\cite{le2017coordinated,yeh2019diverse,ding2020graph,hauri2021multi,omid2022multiagent,alcorn2021baller2vec}. This task involves forecasting future positions and motions of multiple agents in a shared environment. In multi-agent sports like soccer, accurate motion prediction deepens insights into player behavior, team dynamics, and on-field decision-making processes.

\begin{figure*}
  \centering
  \vspace{-2mm}
  \includegraphics[width=1.0\textwidth]{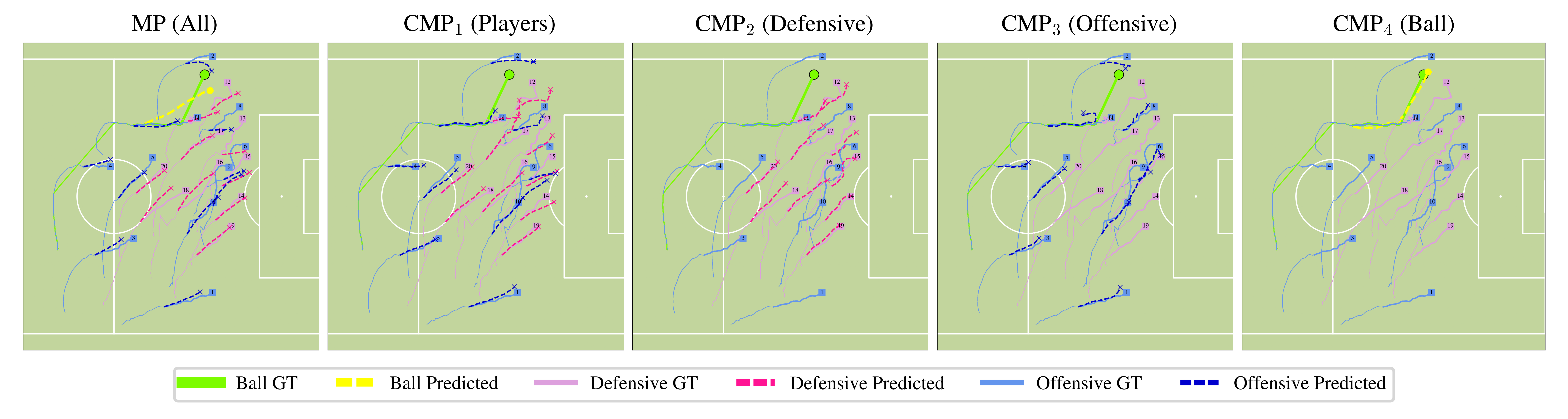}
  \vspace{-0.8cm}
  \caption{\textbf{Motion prediction in soccer.} The method predicts both player and ball motions from partial 2D trajectories under specified conditions. In the figure, squares represent the end positions of ground truth offensive and defensive team players, crosses denote their predicted positions, and circles indicate the final ball ones. Five different tasks (MP, CMP$_{1-4}$) for the same test sequence are displayed, everyone of them is tailored to predict specific subsets of agents, as specified in parentheses.}
  \vspace{-3mm}
  \label{fig:Introduction}
\end{figure*}

The intricacies of soccer, characterized by swift changes in player positions, rapid ball movements, and intricate team coordination, underscore the need for advanced models surpassing conventional non-social-aware motion prediction. In image processing and computer vision, these challenges crucially apply to enhancing player tracking and re-identification, where precise position forecasts contribute to performance analysis, team strategies, and overall gameplay understanding. These forecasts can also be used to simulate team strategies and obtain advanced metrics~\cite{teranishi2022evaluation}. The dynamics of soccer, coupled with player interactions, drive the Conditioned Motion Prediction (CMP) task, adressing scenarios like predicting player trajectories based on ball position~\cite{yeh2019diverse, alcorn2021baller2vec}. Figure~\ref{fig:Introduction} illustrates the MP task and four different CMP tasks in detail that we consider in this paper.

Given the dynamic nature of soccer, a robust model must exhibit permutation equivariance~\cite{yeh2019diverse,ding2020graph}, adapting to varying player compositions and interactions. Transformer-based architectures~\cite{vaswani2017attention}, known for their handling of varying-length sequences and permutation equivariance, have gained traction in motion prediction tasks~\cite{ngiam2021scene,girgis2021latent}, including sports~\cite{alcorn2021baller2vec}.

This research introduces a comprehensive approach to soccer MP and CMP, employing a transformer-based model. Our model captures soccer's intricate properties, leveraging permutation equivariance and historical data to predict ball and player 2D trajectories. Evaluation against baselines showcases social awareness in soccer. The study introduces a synthetic dataset tailored for this research, and utilizes a real one from LaLiga 2022-2023.

\vspace{-2mm}
\section{Related work}
The evolution of multi-agent motion prediction originates from human pose techniques, initially utilizing Recurrent Neural Networks (RNN)~\cite{fragkiadaki2015recurrent, martinez2017human} to capture temporal dynamics. This evolution extended to pedestrian motion, fusing RNN with social pooling~\cite{alahi2016social, gupta2018social} to capture social interactions. However,~\cite{becker2018red} introduced an RNN baseline void of social encoding, yet surpassing social pooling methods. 

Recognizing the significance of social interactions led to the adoption of Graph Neural Networks (GNN) in pedestrian modeling~\cite{kosaraju2019social, salzmann2020trajectron++}, coupled with recurrent techniques. Additionally, siMLPe~\cite{guo2023back} demonstrates the effectiveness of a Multi-Layer Perceptron (MLP) architecture in capturing temporal and spatial dynamics for human motion prediction. Transformer-based architectures also made substantial contributions, showcasing their ability to encode temporal and social dynamics in human pose estimation~\cite{aksan2021spatio} and in forecasting pedestrian and vehicle trajectories~\cite{ngiam2021scene, girgis2021latent, yuan2021agentformer}.

In sports, initial methods focused on generating long-term basketball trajectories using Variational Autoencoders (VAE)~\cite{felsen2018will} and Variational Recurrent Neural Networks (VRNN)~\cite{zhan2018generating, zheng2016generating}.  These variational methods were later outperformed by a multi-modal RNN-based architecture~\cite{hauri2021multi}. To leverage permutation equivariance without the necessity of agent ordering, subsequent research integrated VRNN with GNN for generating multi-agent trajectories in basketball and soccer~\cite{yeh2019diverse, sun2019stochastic, omid2022multiagent}. Addressing concerns associated with accumulated errors in recurrent methods, \cite{ding2020graph} combined Graph Attention Networks (GAT) with temporal convolutional networks in soccer. Transformer-based models have also found applications in the sports domain, demonstrating superior performance compared to graph-recurrent-based approaches~\cite{alcorn2021baller2vec, alcorn2021baller2vec++} in NBA trajectories. Nevertheless, simultaneously conducting attention in both temporal and social dimensions still incurs a notable computational cost.

Our method introduces a tailored transformer encoder-decoder for soccer, adeptly adapting to the sport's intricacies involving a higher number of agents compared to basketball. To enhance computational efficiency, the model is optimized by sequentially decoupling temporal and social attentions. We leverage permutation equivariance alongside the agents' ordering. Moreover, we showcase the effectiveness of our approach in addressing soccer's CMP task using a tailored synthetic dataset and a real one, skillfully capturing intricate agent interactions.

\vspace{-4mm}
\section{Our Method}
\label{ProblemFormulation}

\subsection{Problem statement}
\label{sec:problemformulation}
Consider a set of $M \in \mathbb{N}$ agent measures ($M-1$ players and a ball in our context), $X = \{\mathbf{x}^1, \ldots, \mathbf{x}^M\}$ where every measure contains $k$ elements. The measures evolve over a time horizon of $t+T \in \mathbb{N}$ where $t$ and $T$ are positive integers. Particularly, $t$ represents the observations, while $T$ covers predictions spanning approximately 4 seconds~\cite{le2017coordinated,yeh2019diverse, ding2020graph}. On the one hand, we can define both prior $\mathcal{X}_{0:t}$ and posterior $\mathcal{X}_{t+1:t+T}$ sequences where specific $\mathbf{x}^m_t$ are collected to define our MP problem as:  
\begin{equation}
   f(\mathcal{X}_{0:t})=\mathcal{X}_{t+1:t+T} ,
\end{equation}
where $f$ represents a function to infer the posterior data from the prior. 

On the other hand, we also consider a CMP problem. In particular, CMP predicts the positions of some specific agents ($P$), given the positions of other ones ($C$) in a scenario. Let $\mathcal{X}^{C}_{0:t+T}$ represent the complete sequence of observations for the $C$ agents, and let $\mathcal{X}^{P}_{0:t}$ and $\mathcal{X}^{P}_{t+1:T}$ denote the prior and posterior states of the agents to be predicted, respectively. Then we can define the model $f_c$ to sort out the CMP as:
\begin{equation}
f_c(\mathcal{X}^{C}_{0:t+T}, \mathcal{X}^{P}_{0:t}) = \mathcal{X}^{P}_{t+1:t+T}.
\end{equation}

The challenge in MP is to forecast trajectories of all $M$ agents (players and ball). In contrast, CMP encompasses various tasks within soccer that we introduce next:

\noindent\textbf{CMP$_1$}: Predicting players' positions with the ball as a conditioning agent. 

\noindent\textbf{CMP$_2$}: Predicting defensive team players' positions using all other agents as conditioning ones. 

\noindent\textbf{CMP$_3$}: Predicting offensive team players' positions using all other agents as conditioning ones. 

\noindent\textbf{CMP$_4$}: Predicting the ball position with players as conditioning agents. 

\begin{figure}[t]
  \centering
  \includegraphics[width=0.95\columnwidth]{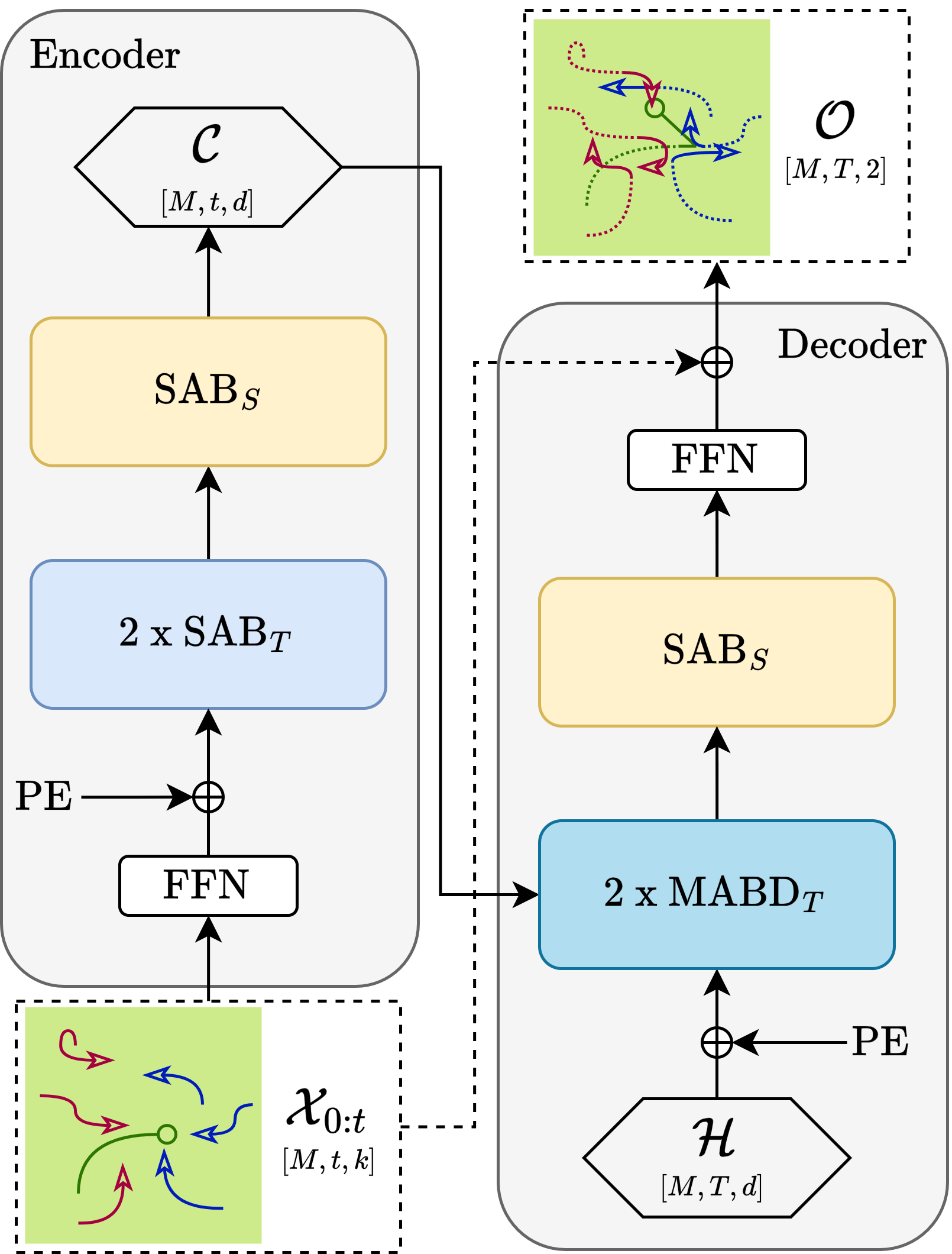}
  \vspace{-4mm}
  \caption{\textbf{FootBots architecture in soccer.} FootBots exploits an encoder-decoder structure with sequential temporal and social attention mechanisms. It incorporates Set Attention Blocks to encode temporal SAB$_T$ and social SAB$_S$ dynamics represented in the context $\mathcal{C}$. The Multi-Attention Block Decoder in the temporal axis (MABD$_T$) and SAB$_S$ in the decoder generate the predicted trajectories. FootBots is capable of solving both MP and CMP tasks in soccer, with an input of the decoder $\mathcal{H}$ varying depending on the task.}
  \label{fig:FootBots}
  \vspace{-5mm}
\end{figure}

\subsection{Attention mechanisms}
Attention mechanisms are effective at capturing relationships in sequences or sets. Given $n$ queries $\mathbf{Q}$ of dimension $d_k$, $n_v$ keys $\mathbf{K}$ of dimension $d_k$, and $n_v$ values $\mathbf{V}$ of dimension $d_v$, the attention mechanism computes weighted value sums using compatibility between queries and keys as:
{\small
\begin{equation}
    \text{Attention}(\mathbf{Q}, \mathbf{K}, \mathbf{V}) = \text{softmax}\left(\frac{\mathbf{Q}\mathbf{K}^{\top}}{\sqrt{d_k}}\right)\mathbf{V},
\end{equation}}
with $\mathbf{Q} \in \mathbb{R}^{n \times d_k}$, $\mathbf{K} \in \mathbb{R}^{n_v \times d_k}$, $\mathbf{V} \in \mathbb{R}^{n_v \times d_v}$. In practice, the attention mechanism is often extended with multiple attention heads, also called Multi-Head Attention (MHA), originally introduced in Transformer architecture~\cite{vaswani2017attention}, allowing the model to capture different relations in the data.

The MHA operation was extended to work on sets by defining a Set Attention Block (SAB)~\cite{lee2019set}, an adaptation of the encoder block of the Transformer that lacks the positional encoding. The MHA itself provides the property of permutation equivariance, making the SAB a permutation-equivariant operation. Finally, the original Transformer~\cite{vaswani2017attention} builds the output using its decoder, also called Multi-Attention Block Decoder (MABD)~\cite{girgis2021latent}, which utilizes cross-attention to take into account the encoder output.

In motion prediction, attention mechanisms can capture temporal dynamics and social interactions among agents. Temporal attention focuses on sequence dynamics, assigning varying weights to observations for accurate future motion prediction. Social attention complements that by considering interactions among agents, capturing spatial relationships, and accounting for collaborative behaviors. 

\subsection{FootBots}
In designing FootBots, we leverage SAB and MABD blocks drawing inspiration from~\cite{girgis2021latent, ngiam2021scene}. FootBots utilizes an encoder-decoder structure with sequential temporal and social attention mechanisms, capturing player-ball dynamics over time. Figure~\ref{fig:FootBots} illustrates its key components.

The encoder of FootBots operates by handling input observations denoted as $\mathcal{X}_{0:t}$ through a Feed-Forward Network ($\mathrm{FFN}$), which is supplemented with a positional encoder ($\mathrm{PE}$) to ensure the temporal ordering of the data. To capture the dynamics in both time and social interactions, we utilize SAB. More specifically, SAB$_T$ for temporal encoding dynamics and SAB$_S$ for social encoding. Those blocks are responsible for capturing interactions among the players and the ball during the prior state, resulting in the generation of a representation tensor called context, denoted as $\mathcal{C}$.

Given the prior sequence of sets $\mathcal{X}_{0:t} = (X_0, \ldots, X_t)$ with dimensions $[M,t,k]$, we define the encoder operations as follows:
\begin{equation}
\small
\mathcal{C} = \text{SAB}_S \left( \text{SAB}_T (\text{SAB}_T (\textrm{PE} + \text{FFN}(\mathcal{X}_{0:t}) )) \right),
\end{equation}
where $\mathcal{C}$ has the dimension $[M,t,d]$, and $d$ is the chosen dimension for the embeddings.

In the decoder, FootBots generates predicted trajectories $\mathcal{O}$ to approximate $\mathcal{X}_{t+1:t+T}$. To this end, it employs a MABD in the temporal axis (MABD$_T$) and a SAB$_S$ in the social one. MABD$_T$ takes into account the output of the encoder $\mathcal{C}$ and the input of the decoder $\mathcal{H}$, which depends on the task at hand: 1) in MP, $\mathcal{H}$ relies on the last $T$ time steps of $\mathcal{C}$; 2) in CMP, $\mathcal{H}$ incorporates the observations of the conditioning agents during the prediction interval $\mathcal{X}^{C}_{t+1:t+T}$, along with the last $T$ time steps of $\mathcal{C}$ for the agents of interest ($P$), $\mathcal{C}^{P}_{t-T:t}$. 

Given $T \le t$, where $T$ represents the desired frames to predict, we outline the decoder operations in the following equation:
\begin{equation} 
\small
\mathcal{O} = \mathcal{X}_{t} + \text{FFN}\left(\text{SAB}_S \left( \text{MABD}_T (\text{MABD}_T (\textrm{PE} + \mathcal{H}, \mathcal{C}), \mathcal{C}) \right) \right),
\end{equation}
where $\mathcal{X}_{t}$ is the last set of observations of the prior and $\mathcal{H}$ is one input of the MABD$_T$ operation whose definition depends on the task: $\mathcal{H} = \mathcal{C}_{t-T:t}$ for MP; and $\mathcal{H} = \text{FFN}(\mathcal{X}^{C}_{t+1:t+T}) \cup \mathcal{C}^{P}_{t-T:t}$ for CMP. It is worth noting that these relations justify our model's constraint that $T \leq t$.  

The dimension of $\mathcal{O}$ is $[M,T,2]$. A skip connection enables residual learning, particularly benefiting early frame precision. In MP task, FootBots maintains agent permutation equivariance, while in CMP task, it demonstrates partial equivariance for both conditioning agents ($C$) and agents of interest ($P$), preserving this property within their respective subsets.

For completeness, we also propose a non-social variant of our approach denoted by FootBots NS. In this case, our method substitutes social SAB$_S$ attention with temporal SAB$_T$ one, omitting social interactions to showcase their importance. 

\subsection{Loss}

The loss function utilized is an Average Displacement Error (ADE), a widely used loss metric in trajectory prediction tasks. It computes the average $l_2$-norm between the predicted trajectories and the ground truth (GT) ones as: 
\begin{equation}
\label{eq:ADE}
    \text{ADE} = \frac{1}{M T} \sum_{m=1}^{M} \sum_{j=t+1}^{t+T} \left\lVert \hat{\mathbf{x}}^{m}_{j} - \mathbf{x}^{m}_{j} \right\rVert_2,
\end{equation}
where $\hat{\mathbf{x}}^{m}_{j}$ and $\mathbf{x}^{m}_{j}$ represent the predicted and its corresponding GT position of the $m$-th agent at $j$-th time step.

\section{Experimental evaluation}
In this section, we present our experimental results on motion prediction and provide a comparison with respect to competing approaches. For quantitative evaluation, we consider a subset of agents $\hat{M}$ by using four types of metrics. First, we consider the $\textrm{ADE}_{\hat{M}}$ metric in Eq.~\eqref{eq:ADE} for just the set $\hat{M}$. 

Second, we propose a Final Displacement Error (FDE) that measures the final deviation between the prediction and the corresponding ground truth location as:
\begin{equation} \nonumber
\small
\textrm{FDE}_{\hat{M}} = \frac{1}{|\hat{M}|} \sum_{m \in \hat{M}} \left\lVert \mathbf{x}^{m}_{t+T} - \hat{\mathbf{x}}^{m}_{t+T} \right\rVert_2 \, .
\end{equation} 

For completeness, we also propose to compute the Maximum Error (MaxErr) as:
\begin{equation} \nonumber
\small
\textrm{MaxErr}_{\hat{M}} = \frac{1}{|\hat{M}|} \sum_{m \in \hat{M}} \max_{j \in \{t+1, \ldots, t+T\}} \left\lVert \mathbf{x}^{m}_{j} - \hat{\mathbf{x}}^{m}_{j} \right\rVert_2 \, ,
\end{equation} 
and the missing rate (MR) to show the percentage of predictions having an $l_2$-norm greater than 1 meter as:
\begin{equation} \nonumber
\small
\textrm{MR}_{\hat{M}} = \frac{1}{|\hat{M}| T} \sum_{m \in \hat{M}} \sum_{j=t+1}^{t+T} \mathbb{I}\left[\left\lVert \mathbf{x}^{m}_{j} - \hat{\mathbf{x}}^{m}_{j} \right\rVert_2 > 1\right],
\end{equation}
with $\mathbb{I}(\cdot)$ an indicator function.

\subsection{Datasets}
In this paper, we propose to validate our model on synthetic and real datasets. Next, we provide the most important details for each of them. 

\noindent\textbf{Synthetic dataset}: Created to aid investigation and model development for both MP and CMP tasks, this dataset contains 10,000 training and 1,500 validation sequences. Each sequence spans 20 time steps: 10 for prior ($t$) and the rest for target ($T$). Five agents, including four players and one ball, compose each sequence. The ball follows a linear trajectory initially, but can randomly change direction at a chosen time step, followed by another linear path. Noise is introduced for trajectory variability, causing slight deviations from linearity. Player behaviors encompass remaining stationary with noise (S), linear trajectories with noise (L), and non-linear paths influenced by the ball's position as an attractor (A). The number of players for each behavior is randomly determined, all within a bounded square region $[-15, 15] \times [-15, 15]$ resembling meters in real world. In this dataset, each agent's observation is limited to its $xy$ location, leading to $k=2$ according to the problem formulation.

\noindent\textbf{Real dataset:} 
This dataset comprises actual data from 283 matches of LaLiga's 2022-2023 season, capturing agent motions using advanced computer vision techniques. Each match is divided into sequences representing 9.6 seconds, down-sampled to 6.25 frames per second. Each sequence, consisting of 60 frames, is divided into prior states (35 frames or 5.6 seconds) and target ones (25 frames or 4 seconds). Only trajectories of all 20 field players (excluding goalkeepers) are considered, and agent order is standardized. The dataset is split into 243 matches (82,954 sequences) for training, 20 matches (6,258 sequences) for testing, and 20 matches (7,500 sequences) for validation. Trajectories are normalized to fit within $[-1,1] \times [-1,1]$ by dividing by the largest pitch dimension, and spatial realignment ensures the possession team's rightward motion on the pitch. When using FootBots and its non-social variant FootBots NS, each agent's observation includes its 2D position and an associated integer indicating its role: ball, defensive team player, or offensive team player. Therefore, in this case we consider $k=3$ elements. In contrast, when using other baseline models, input is confined to the 2D position, leading to $k=2$.

\subsection{Synthetic evaluation}
This initial scenario aims to emphasize the motivation and importance of effectively solving the CMP tasks in soccer. By analyzing the synthetic dataset, we can evaluate the significance of the social SAB$_S$ and its ability to address the specific CMP$_1$ task.

The outcomes of our proposed methods on the synthetic dataset are detailed in Table~\ref{tab:exp2synthetic}, highlighting the performance metrics for both MP and CMP$_1$ tasks. In terms of $\textrm{ADE}_{P}$, FootBots exhibit a slight advantage over FootBots NS when addressing the MP task. This divergence can be attributed to FootBots' capacity to anticipate the motions of ball-attracted players (A) by leveraging the anticipated ball position, facilitated by the social SAB$_S$. However, it is important to note that despite these enhancements, the persistence of high $\textrm{ADE}_{\text{ball}}$ values implies that predictions for type A players may be subject to error propagation. Shifting to the CMP$_1$ task, a notable improvement in predictions is observed. 

\begin{table}[t]
  \centering
  \footnotesize
  \begin{tabular}{@{}lccccc@{}}
    \toprule
    \text{Model} 
    & \text{Task}
    & Predict ($P$)
    & $\textrm{ADE}_P$(m) $\downarrow$
    & $\textrm{ADE}_\text{ball}$(m) $\downarrow$ \\
    \midrule
    FootBots NS & MP & Players+Ball & 0.50 & 1.83 \\
    FootBots & MP & Players+Ball & 0.44 & 1.72 \\
    FootBots & CMP$_1$ & Players & \textbf{0.10} & -  \\
    \bottomrule
  \end{tabular}
  \vspace{-2mm}
  \caption{\textbf{Evaluating our architecture on synthetic data.}}
  \label{tab:exp2synthetic}
  \vspace{-3mm}
\end{table}

\begin{figure}[t]
  \centering
  \includegraphics[trim={0 0 0 2.5cm},clip, width=1.0\columnwidth]{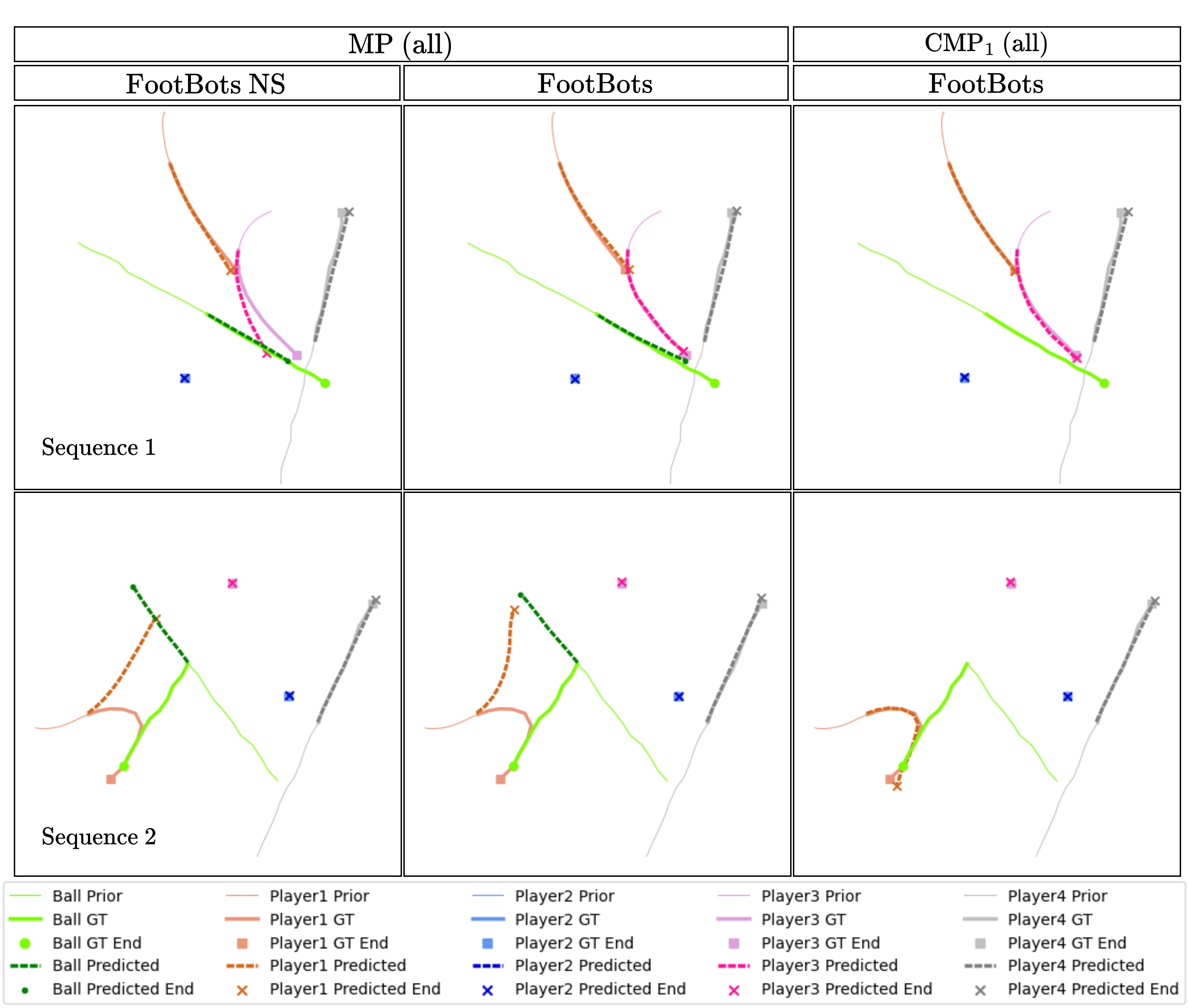}
  \vspace{-8mm}
  \caption{\textbf{Two examples from the synthetic dataset.} The examples serve to visually compare the performance of FootBots NS and FootBots solving MP task, and FootBots solving CMP$_1$. The predictions for different player types (S, L, and A) are evaluated, emphasizing the impact of incorporating social attention and the ball as the conditioning agent.}
  \label{fig:synthetic_FootBots}
  \vspace{-5mm}
\end{figure}

Nevertheless, for a deeper evaluation, qualitative analysis of example sequences is crucial. Figure~\ref{fig:synthetic_FootBots} provides further insight, showcasing two instances from the validation set, each featuring distinct complexities in ball trajectory. These instances enable differentiation between FootBots NS and FootBots in the MP task, as well as between FootBots in MP and FootBots in CMP$_1$. All three models adeptly predict static (type S) and linear (type L) player positions with commendable accuracy. However, FootBots NS faces challenges in accurately predicting type A player actions due to its reliance on extrapolating their past trajectories without considering ball-related factors. In the context of the MP task with FootBots, a noteworthy correlation emerges between predictions for type A players and the quality of ball prediction. Consequently, the precision of type A predictions is significantly reliant on accurate ball prediction. This is exemplified in sequence 1, where the ball trajectory retains linearity throughout the prediction time-frame, leading to almost precise predictions. Conversely, in sequence 2, deviations in ball prediction propagate errors to the predictions of type A players. The robustness of our model's capacity to effectively address CMP$_1$ task (conditioned on ball information), is emphasized in the concluding segments of our study. Across the two scenarios presented, the model consistently achieves nearly precise predictions within the CMP$_1$ context.

\vspace{-2mm}
\subsection{Real evaluation}
\label{subsec:RealEval}

\begin{figure*}[t!]
  \centering
  \includegraphics[trim={0 0 0 3mm},clip, width=1.0\linewidth]{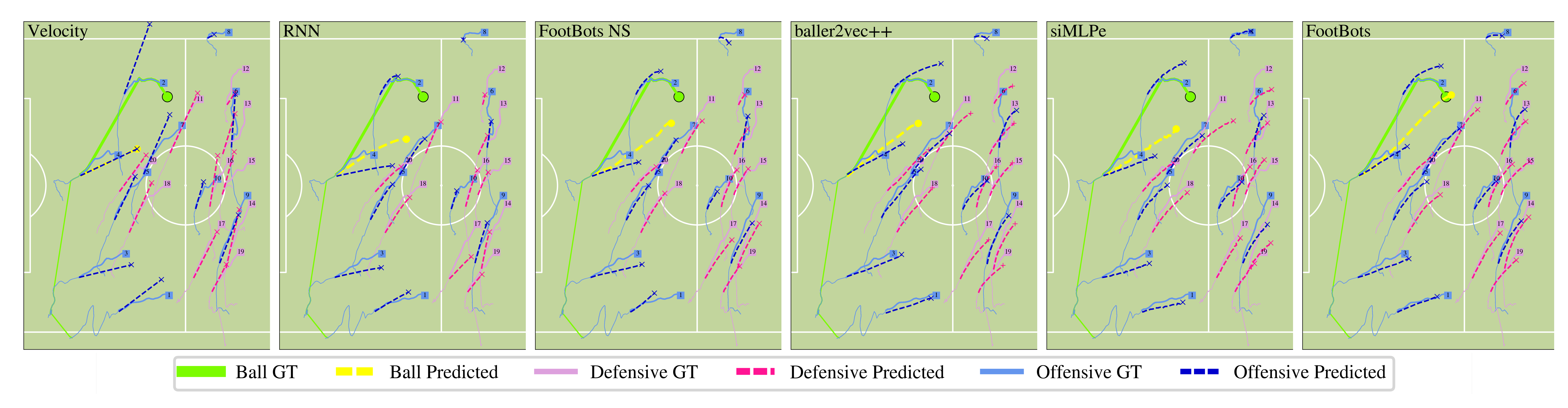}
  \vspace{-10mm}
  \caption{\textbf{Qualitative evaluation and comparison on real data.} The figure displays the estimated trajectories for approaches Velocity, RNN~\cite{becker2018red}, baller2vec++~\cite{alcorn2021baller2vec++} and siMLPe~\cite{guo2023back}; and our solutions FootBots NS and FootBots, by solving the MP task.}
  \label{fig:Exp1_Seq105}
  \vspace{-2mm}
\end{figure*}

\begin{table*}[t!]
  \centering
  \small
  \begin{tabular}{lcccccccc}
    \toprule
    \text{Model} & \text{Order} & Task & Predict ($P$) & $\textrm{ADE}_P$ $\downarrow$  & $\textrm{ADE}_\text{ball}$ $\downarrow$ & $\textrm{MaxErr}_P$ $\downarrow$ & $\textrm{FDE}_P$ $\downarrow$ & $\textrm{MR}_P$ (\%) $\downarrow$ \\
    \midrule
    Velocity & None & MP & Players+Ball & 3.27 & 9.39 & 7.34 & 7.27 & 67.50 \\
    RNN~\cite{becker2018red} & None & MP & Players+Ball & 2.67 & 6.91 & 5.56 & 5.43 & 65.54 \\
    FootBots NS (Ours) & None & MP & Players+Ball & 2.39 & 6.37 & 5.16 & 5.04 & 60.99\\
    baller2vec++~\cite{alcorn2021baller2vec++} & Approx-Equivariant & MP & Players+Ball & 2.21 & 6.43 & \underline{4.64} & \underline{4.49} & 60.79\\
    siMLPe~\cite{guo2023back} & Role-based & MP & Players+Ball & \underline{2.18} & \underline{6.15} & 4.73 & 4.55 & \underline{59.71} \\
    \midrule
    \multirow{5}{*}{FootBots (Ours)} & \multirow{5}{*}{Equivariant} & MP & Players+Ball & \textbf{2.04} & \textbf{5.79} & \textbf{4.43} & \textbf{4.28} & \textbf{57.37} \\
    & & CMP$_1$ & Players & 1.64 & - & 3.42 & 3.20 & 52.59 \\
    & & CMP$_2$ & Defensive & 1.38 & - & 2.80 & 2.59 & 47.83 \\
    & & CMP$_3$ & Offensive & 1.44 & - & 2.98 & 2.78 & 48.26 \\
    & & CMP$_4$ & Ball & 2.72 & 2.72 & 5.93 & 4.27 & 64.11 \\
    \bottomrule
  \end{tabular}
  \vspace{-2mm}
  \caption{\textbf{Quantitative evaluation and comparison for MP and CMP tasks on real data.} The table provides a comprehensive comparison of our solution with various other approaches in MP task. All metrics, except $\textrm{MR}_P$, are in meters.}
  \label{tab:exp1}
  \vspace{-5mm}
\end{table*}

In the second evaluation scenario, we analyze the efficacy of FootBots in addressing the MP task by employing a real dataset. We conduct a comparative assessment with various baseline methods to gauge its performance. Additionally, utilizing the same real dataset, we evaluate FootBots' performance in the CMP tasks. The considered baselines, including the already described \textbf{FootBots NS}, are outlined as follows:

\noindent\textbf{Velocity}: We employ velocity extrapolation as a preliminary benchmark, projecting agent predictions linearly based on observed velocity. 

\noindent\textbf{RNN}: We implement an RNN with LSTM cells, using an encoder for input representation and an MLP decoder for prediction, a method proven effective in prior work~\cite{becker2018red}.

\noindent\textbf{siMLPe}: Adapted from human pose prediction, siMLPe treats each joint as an agent, utilizing an MLP-based model with layer normalization and transposition operations for spatial-temporal dynamics using a fixed sequence length~\cite{guo2023back}. 

\noindent\textbf{baller2vec++}: Adapted from the basketball context, this approach conducts attention simultaneously in both temporal and social dimensions by modelling the attention mask~\cite{alcorn2021baller2vec++}.

It is important to note that Velocity, RNN~\cite{becker2018red}, and FootBots NS operate independently for each agent, making the order of the input irrelevant. However, siMLPe~\cite{guo2023back} is a role-based model and not equivariant. In baller2vec++, they demonstrate minimal result variation when changing the ordering of agents, describing it as approximately equivariant. To ensure a fair comparison, the real dataset is ordered based on the initial positional role of each player.

In Table~\ref{tab:exp1}, we provide a comprehensive overview of metrics across methods solving the MP task. Our FootBots demonstrate superior performance in all metrics. The Velocity model shows a significant performance gap, attributed to unconstrained long-term predictions exceeding pitch boundaries. RNN~\cite{becker2018red} and FootBots NS, lacking agent interaction, lead to performance decline, mostly in the $\textrm{MaxErr}_{P}$ and FDE$_{P}$ metrics. This emphasizes the significance of social interaction in long-term trajectory prediction. In general, social-aware baselines like siMLPe~\cite{guo2023back} and baller2vec++~\cite{alcorn2021baller2vec++} achieve superior metrics compared to non-social methods. Although siMLPe competes well, $\textrm{MaxErr}_{P}$ and FDE$_{P}$ metrics are outperformed by transformer-based approaches like baller2vec++ and FootBots. Moreover, these methods can handle variable sequence length. However, baller2vec++ encounters difficulties with ball prediction, leading to suboptimal results due to error accumulation. FootBots excels in ball and players predictions, leverages permutation equivariance, and is more than six times faster than baller2vec++ in inference (73 vs 484 milliseconds), thanks to the decoupled attention. Importantly, in all MP results, $\textrm{ADE}_{ball}$ consistently exceeds $\textrm{ADE}_{P}$ across methods, motivating CMP$_1$ task.

Figure~\ref{fig:Exp1_Seq105} provides an illustration of a particular trajectory prediction example, offering a comparative analysis of baselines in the MP task. Linear predictions based on Velocity baseline exhibit trajectories that are deemed non-sensical in certain instances. Both FootBots NS and RNN~\cite{becker2018red} models tend to generate shorter predicted trajectories, underscoring the imperative of incorporating social interaction for a comprehensive understanding of each player's future motions. Despite suboptimal ball prediction, siMLPe~\cite{guo2023back} and baller2vec++~\cite{alcorn2021baller2vec++} excels in accurate player predictions, showcasing its robustness in capturing player interaction dynamics. Additionally, FootBots outperforms previous baselines both quantitatively and qualitatively, with enhanced ball prediction.

To ensure the generalization of our findings, we conduct a parallel analysis using the real dataset to address all the considered CMP tasks. Similar to the synthetic dataset, we initiate the evaluation by assessing the model's performance in solving CMP$_1$. Our investigation also covers CMP$_2$, focusing on Defensive Players' positions, and CMP$_3$, targeting Offensive Players' ones. Furthermore, we explore CMP$_4$, which involves ball position prediction.

Quantitative results for FootBots across all CMP tasks are presented in Table~\ref{tab:exp1}. The solution for the CMP$_1$ task demonstrates marked improvement compared to the MP task, highlighting a strong correlation between player positions and the ball. Noteworthy enhancements in prediction accuracy emerge when conditioning on the opposing team and ball locations in CMP$_2$ and CMP$_3$. It is worth noting that predicting offensive team behaviors is more challenging than predicting defensive ones, due to their inherent stochasticity. Moreover, FootBots adeptly utilize player interactions to provide accurate ball predictions, reducing the $\textrm{ADE}_\text{ball}$ metric from 5.79 to 2.72 meters compared to the MP task.

Figure~\ref{fig:Introduction} illustrates a sample of the real dataset featuring solutions for the MP task and all CMP tasks. This specific instance depicts an scenario characterized by an extended ball trajectory involving a pass to predict and swift player motions. Notably, within the MP task, the model encounters challenges in accurately predicting both the ball and player positions, attributed to the inherent speed of the sequence. However, with the integration of conditioning information, discernible enhancements in predictions become evident.

\vspace{-2mm}
\section{Conclusions}
In this work, we introduced FootBots, a tailored trajectory prediction model for soccer contexts, and extensively evaluated its performance across diverse scenarios. The comparative analysis demonstrated FootBots' superior performance over baseline models, showcasing its advantageous equivariance properties. Through synthetic dataset evaluation, FootBots excelled in predicting player positions, particularly when incorporating social attention and ball conditioning (CMP$_1$), highlighting the importance of social interactions and ball incorporation. Extension to real data showcased FootBots' grasp of defensive strategies (CMP$_2$), improved offensive player predictions (CMP$_3$), and effective player interaction utilization for ball position enhancement (CMP$_4$). Remarkably, CMP$_4$ exhibited significant error reduction compared to the MP task, affirming the effectiveness of using players as conditioning agents to enhance ball prediction accuracy.

\bibliographystyle{IEEEbib}
\bibliography{FootBots_ICIP}

\end{document}